\documentclass[runningheads]{files/llncs}
\usepackage{xfrac, amsfonts, amsbsy, algorithm, color, multirow, enumitem, graphicx, amsmath, bm, booktabs}
\usepackage{caption}
\usepackage{subcaption}
\usepackage[noend]{algpseudocode}
\usepackage[section]{placeins}
\usepackage{wrapfig}
\usepackage{times}
\usepackage{soul}

\DeclareMathAlphabet{\pazocal}{OMS}{zplm}{m}{n}

\def \r {\mathbf{r}}
\def \c {\mathbf{c}}
\def \o {\mathbf{o}}
\def \d {\mathbf{d}}
\def \w {\mathbf{w}}
\def \T {\mathbf{T}}
\def \K {\mathbf{K}}

\def\y {\mathbf{y}}

\def\d {\mathbf{d}}

\def\X {\mathbf{X}}

\def\w {\mathbf{w}}

\def \d {\mathbf{d}}

\def \Th {\pmb{\Theta}}

\def\Th{\mathbf{\Theta}}

\def \L {\pazocal{L} }

\def \R {\mathcal{R}}

\def \Rspace { \mathbb{R}}

\def\argmax{\mathop{\mathrm{arg\,max}}}

\begin{document}

\title{Learning Neural Radiance Fields of Forest Structure for Scalable and Fine Monitoring}

\author{Juan Castorena\inst{1}} 
\authorrunning{J. Castorena}
\institute{Los Alamos National Laboratory, Los Alamos, NM, USA, 48124 \\
\email{jcastorena@lanl.gov}}

\maketitle
\begin{abstract}
    This work leverages neural radiance fields and remote sensing for forestry applications. Here, we show neural radiance fields offer a wide range of possibilities to improve upon existing remote sensing methods in forest monitoring. We present experiments that demonstrate their potential to: (1) express fine features of forest 3D structure, (2) fuse available remote sensing modalities and (3), improve upon 3D structure derived forest metrics. Altogether, these properties make neural fields an attractive computational tool with great potential to further advance the scalability and accuracy of forest monitoring programs. 
\end{abstract}

\section{Introduction}

With approximately four billion hectares covering around $~31 \%$ of the Earth's land area \cite{FaoUnep2020}, forests play a vital role in our ecosystem. The increasing demand for tools that help maintain a balanced and healthy forest ecosystem is challenging due to the complex nature of various factors, including resilience against disease and fire, as well as overall forest health and biodiversity \cite{white2016remote}. Active research focuses on the development of monitoring methods that synergistically collect comprehensive information about forest ecosystems and utilize it to analyze and generate predictive models of the characterizing factors. These methods should ideally be capable of effectively and efficiently cope with the dynamic changes over time and heterogeneity. The goal is to provide the tools with such properties for improved planning, management, analysis, and more effective decision-making processes  \cite{Atchley2021}. Traditional tools for forest monitoring, such as national forest inventory (NFI) plots, utilize spatial sampling and estimation techniques to quantify forest cover, growing stock volume, biomass, carbon balance, and various tree metrics (e.g., diameter at breast height, crown width, height) \cite{tomppo2010national}. However, these surveying methods consist of manual field sampling, which tends to introduce bias and poses challenges in terms of reproducibility. Moreover, this approach is economically costly and time-consuming, especially when dealing with large spatial extents.

Recent advancements, driven by the integration of remote sensing, geographic information and modern computational methods, have contributed to the development of more efficient, cost/time effective, and reproducible ecosystem characterizations. These advancements have unveiled the potential of highly refined and detailed models of 3D forest structure. Traditionally, the metrics collected through standard forest inventory plot surveys have been utilized as critical inputs in applications in forest health \cite{lausch2017understanding}, wood harvesting \cite{kankare2014estimation}, habitat monitoring \cite{vierling2008lidar}, and fire modeling \cite{linn2002}. The efficacy of these metrics relies in their ability to quantitatively represent the full forest's 3D structure including its vertical resolution: from the ground, sub-canopy to the canopy structure. 
%
Among the most popular remote sensing techniques, airborne LiDAR scanning (ALS) has gained widespread interest due to its ability to rapidly collect precise 3D structural information over large regional extents \cite{dubayah2000lidar}. 
Airborne LiDAR, equipped with accurate position sensors like RTK (Real-Time Kinematic), enables large-scale mapping from high altitudes at spatial resolutions ranging from 5-20 points per square meter. It has proven effective in retrieving important factors in forest inventory plots \cite{hyyppa2012advances}. 
However, it faces challenges in dense areas where the tree canopy obstructs the LiDAR signal, even with its advanced full-waveform-based technology.
\textit{In-situ} terrestrial laser scanning (TLS) on the other hand provides detailed vertical 3D resolution from the ground, sub-canopy and canopy structure informing about individual trees,  shrubs, ground surface, and near-ground vegetation at even higher spatial resolutions \cite{hilker2010comparing}. 
Recent work by \cite{pokswinski2021simplified} has demonstrated the efficiency and efficacy of ecosystem monitoring using single scan in-situ TLS. The technological advances of such models include new capabilities for rapidly extracting highly detailed quantifiable predictions of vegetation attributes and treatment effects in near surface, sub-canopy and canopy composition. 
However, these models have only been deployed across spatial domains of a few tens of meters in radius due to the existing inherited limitations of TLS spatial coverage \cite{pokswinski2021simplified}. On the other side of the spectrum, image based photogrammetry for 3D structure extraction offers the potential of being both scalable and the most cost efficient. Existing computational methods for the extraction of 3D structure in forest ecosystems, however, have not been as efficient. Aerial photogrammetry methods result in 3D structure that contains very limited structural information along the vertical dimension and have encountered output spatial resolutions that can be at most only on par with those from ALS \cite{white2016remote}. 

Our contribution seeks to fuse the experimental findings across remote sensing domains in forestry; from broad-scale to in-situ sensing sources. The goal is the ability to achieve the performance quality of \textit{in-situ} sources (e.g., TLS) in the extraction of 3D forest structure at the scalability of broad sources (e.g., ALS, aerial-imagery). We propose the use of neural radiance field (NERF) representations \cite{mildenhall:2020} which account for the origin and direction of radiance to determine highly detailed 3D structure via view-consistency. We observe that such representations enable both the fine description of forest 3D structure and also the fusion of multi-view multi-modal sensing sources. Demonstrated experiments on real multi-view RGB imagery, ALS and TLS validate the fine resolution capabilities of such representations as applied to forests. In addition, the performance found in our experiments of 3D structure derived forest factor metrics demonstrate the potential of neural fields to improve upon the existing forest monitoring programs. To the best of our knowledge, the demonstrations conducted in this research, namely, the application of neural fields for 3D sensing in forestry, is novel and has not been shown previously. In the following, Sec. \ref{Sec:background} provides a brief overview of neural fields. Sec. \ref{Sec:representation} includes experiments illustrating the feasibility of neural fields to represent fine 3D structure of forestry while Section \ref{Sec:fusion} demonstrates the effectiveness of fusing NERF with LiDAR data by enforcing LiDAR point cloud priors. Finally, Section \ref{Sec:prediction} presents results that show the efficacy of NERF extracted 3D structure for deriving forest factor metrics, which are of prime significance to forest managers for monitoring.


\section{Background} \label{Sec:background}
\subsection{Neural Radiance Fields}
The idea of neural radiance fields (NERF) is based on classical ray tracing of volume densities \cite{kajiya:1984}. Under this framework, each pixel comprising an image is represented by a ray of light casted onto the scene. The ray of light is described by $\r(t) = \o + t \d$  with origin $\o \in \Rspace^3$, unit $\ell_2$-norm length direction $\d \in \Rspace^3$ (i.e., $\|\d\|_{2} = 1$)  and independent variable $t \in \Rspace$ representing a relative 
distance. The parameters of each ray can be computed through the camera intrinsic matrix $\K$ with inverse $\K^{-1}$, the 6D pose transformation matrix $\T_{m \rightarrow 0}$ of image $m$ as in Eq. \eqref{ray}
	\begin{eqnarray} \label{ray}
	\nonumber (\o , \d )= \left ( T^{(4)}_{m \rightarrow 0}, \frac{\d'}{ \| \d'\|_{\ell_2}} \right ) \quad \text{with } \\ 
	\d' = \T^{-1}_{m \rightarrow 0} \K^{-1}
	\begin{bmatrix}
	u' \\ v' \\ 1
	\end{bmatrix} - T^{(4)}_{m \rightarrow 0}
	\end{eqnarray}
where $u',v'$ are vertical and horizontal the pixel locations within the image and the subscript $^{(i)}$ denotes the $i$-th column of a matrix. Casting rays $\r \in \R$ into the scene from all pixels across all multi-view images provides information of intersecting rays that can be exploited to infer 3D scene structure. 
Such information consists on sampling along a ray at distance samples $\{t_i\}_{i=1}^M$ and determine at each sample if the color $\c_i \in [0,..,255]^3$ of the ray coincides with those from overlapping rays. If it does not coincide then it is likely that the medium found at that specific distance sample is transparent whereas the opposite means an opaque medium is present. With such information, compositing color can be expressed as a function of ray $\r$ as in Eq. \eqref{model} by:
\begin{equation} \label{model}
\hat{C}(\r) = \sum_{i=1}^{N} \left [   \underbrace{ \left ( \prod_{j=1}^{i-1}  \exp (-\sigma_j \delta_j )  \right ) }_{\text{transparency so far} }  \underbrace{(1-\exp(-\sigma_i \delta_i) )}_{\text{opacity}} \c_i \right]
\end{equation}
where  $\sigma_i \in \Rspace$ and $\delta_i = t_{i+1} - t_i$ are the volume densities and differential time steps  at sample indexed by $i$, respectively. In Eq. \eqref{model} the first term in the summation represents the transparent samples so far while the second term is an opaque medium of color $\c_i$ present at sample $i$.  
Reconstructing a scene in 3D can then be posed as the problem of finding the sample locations $t_i$ where each ray intersects an opaque medium (i.e., where each ray stops ) for all rays casted into the scene. Those intersections are likely to occur at the sample locations where the volume densities are maximized; in other words, where $t_i = \argmax_{i} \{ \mathbf{\sigma} \}$.  Accumulating, all rays casted into the scene and estimating the locations $t_i$'s where volume density is maximized overall rays, renders the 3D geometry of the scene. The number of rays required per scene is an open question; the interested reader can go to \cite{castorena:2010} where a similar problem but for LiDAR sensing determines the number of pulses required for 3D reconstruction depending on a quantifiable measure of scene complexity. 

The problem in Eq. \eqref{model} is solved by learning the volume densities that best explains image pixel color in a 3D consistent way. Learning can be done through a multilayer perceptron (MLP) by rewriting Eq. \eqref{model} as in Eq. \eqref{simplified_model} as: 
\begin{equation} \label{simplified_model}
\hat{C}(\r) = \sum_{i=1}^{N} \w_i \c_i
\end{equation}
where the weights $\w \in \Rspace^N$ encode transparency or opacity of the $N$ samples along a ray and $\c_i$ is its associated pixel color. Learning weights is performed in an unsupervised fashion through the optimization of a loss function using a training set of $M$ pairs of multi-view RGB images and its corresponding 6D poses $\{(\y_m, \T_m)\}_{m=1}^M$, respectively. This loss function
$f : \Rspace^{L} \rightarrow \Rspace$ is the average $\ell_2$-norm error between ground truth color and estimation by compositing described as in Eq. \eqref{loss}:
\begin{equation} \label{loss}
\L_{C} (\Th) = \sum_{\r \in \R} \left [  \|  C( \r) - \hat{C} (\r, \Th)\|_{\ell_2}^2\right]
\end{equation}
Optimization by back-propagation yields the weights that gradually improves upon the estimation of the volume densities.  Other important parameters of NERF are distance $\hat{z}(\r)$ which can be defined using the same weights from Eq.\eqref{model} but here expressed in terms of distance as:
\begin{equation} \label{depth}
	\hat{z}(\r) = \sum_{i=1}^{N}  \omega_i t_i, \quad \qquad  \hat{s}(\r)^2 = \sum_{i=1}^{N}  \omega_i ( t_i - \hat{z}(\r))^2
\end{equation}
and $\hat{s}(\r)$ defined as the standard deviation of distance.
One key issue affecting 3D reconstruction resolution is on the way samples $\{t_i\}_{i=1}^N$ for each ray $\r \in \R$ are drawn. A small number of samples $N$ results in low resolution and erroneous ray intersection estimations while sampling vastly results in much higher computational complexities. To balance this trade-off, the work in \cite{mildenhall:2020} uses two networks one at low-resolution to coarsely sample the 3D scene and another fine-resolution one used subsequently to more finely sample only at locations likely containing the scene.

\section{Are neural fields capable of extracting  3D structure in forestry?}
\label{Sec:representation}
The high capacity of deep learning (DL) models to express data distributions with high fidelity and diversity offers a promising avenue to model heterogenous 3D forest structures in fine detail. The specific configuration of the selected DL model aims to provide a representation that naturally allows the combination of data from multiple sensing modalities and view-points.
Neural fields \cite{mildenhall:2020} under the DL rubric have proven to be a highly effective computational approach for addressing such problems. However, their application has been only demonstrated for indoor and urban environments.

\subsection{ Terrestrial Imagery.} \label{Ssec:individual_tree} 
\begin{figure}[h]
	\centering
		\begin{subfigure}[b]{\textwidth}
		\captionsetup{justification=centering}
		\centering
		\includegraphics[width=0.15\textwidth]{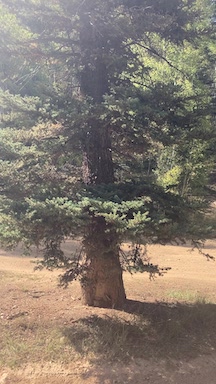}
            \hfill
            \includegraphics[width=0.15\textwidth]{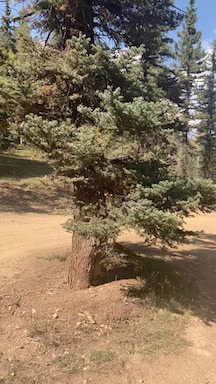}
            \hfill
            \includegraphics[width=0.15\textwidth]{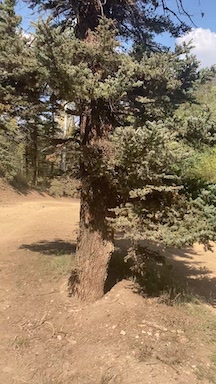}
            \hfill
            \includegraphics[width=0.15\textwidth]{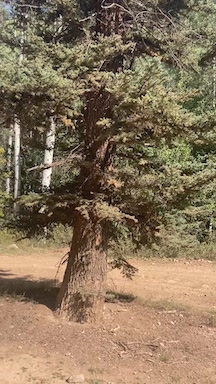}
            \hfill
            \includegraphics[width=0.15\textwidth]{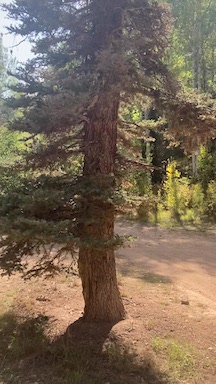}
            \hfill
            \includegraphics[width=0.15\textwidth]{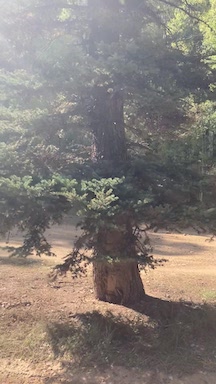}
		\caption{Terrestrial RGB multi-view imagery of Ponderosa Pine Tree.}
		\label{fig:ex1_view1}
	\end{subfigure}

	\begin{subfigure}[b]{0.49\textwidth}
		\centering
		\includegraphics[width=\textwidth]{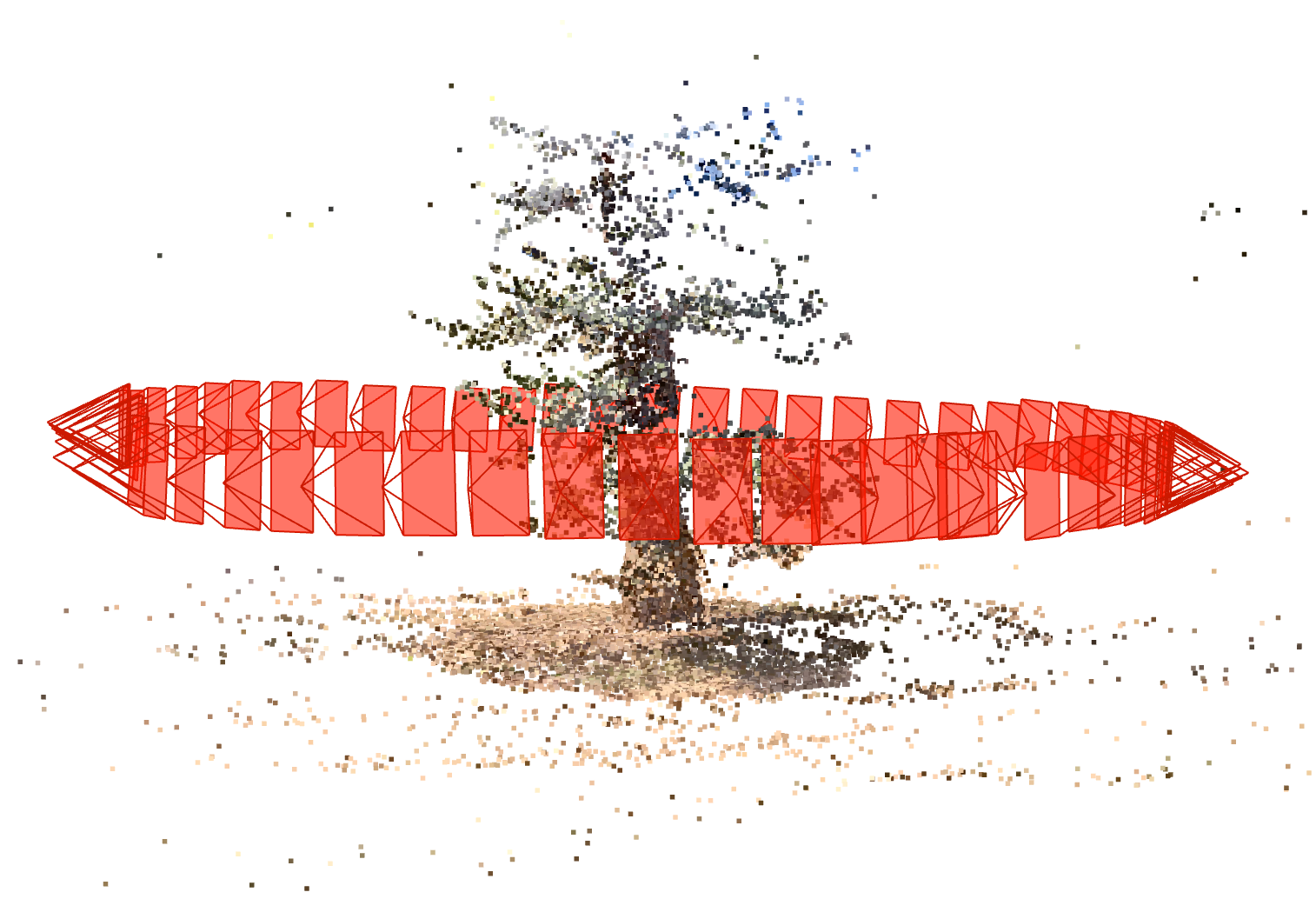}
		\caption{SFM reconstruction view-1}
		\label{fig:ex1_colmap1}
	\end{subfigure}
	\begin{subfigure}[b]{0.49\textwidth}
		\centering
		\includegraphics[width=\textwidth]{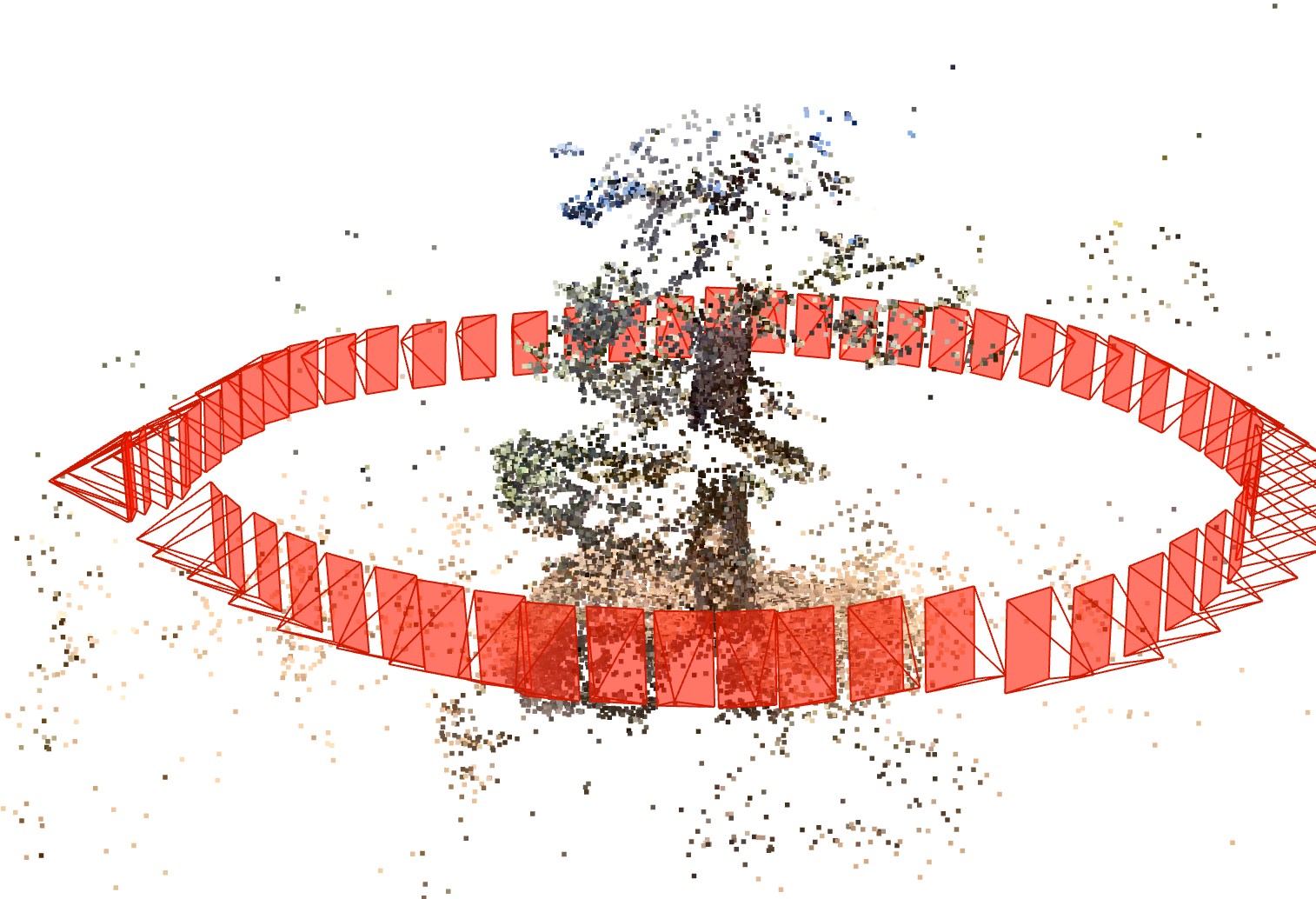}
		\caption{SFM reconstruction view-2}
		\label{fig:ex1_colmap2}
	\end{subfigure}
	
	\caption{Even though SFM reconstruction is capable of extracting the 3D structure of tree, its recontruction suffers from sparsity. Such sparsity limits the spatial variability of structure that can be captured thorugh such models.}
	\label{fig:it_colmap}
\end{figure}

\begin{figure}[h]
	\centering

     \begin{subfigure}[b]{\textwidth}
		\centering
		\includegraphics[width=0.22\textwidth]{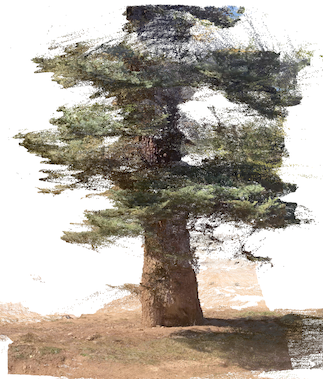}
            \hfill
            \includegraphics[width=0.24\textwidth]{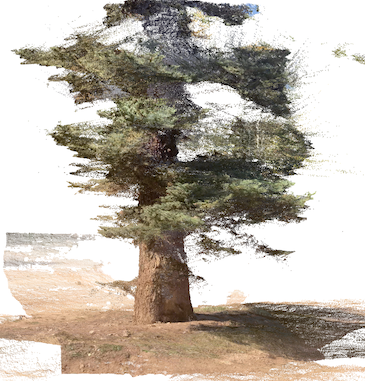}
            \hfill
            \includegraphics[width=0.25\textwidth]{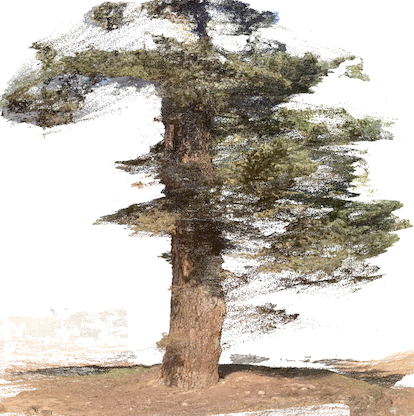}
            \hfill
            \includegraphics[width=0.23\textwidth]{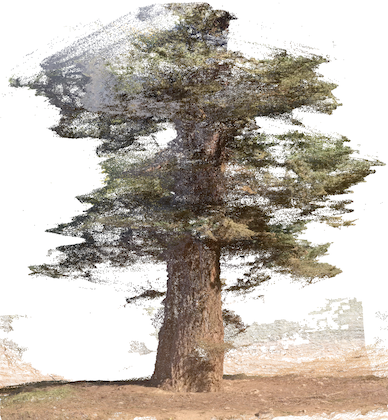}
		\caption{Side views illustrating high 3D spatial detail along the vertical tree stem}
		\label{fig:ex_nerf1}
    \end{subfigure} 



	\begin{subfigure}[b]{0.365\textwidth}
	\centering
	\includegraphics[width=\textwidth]{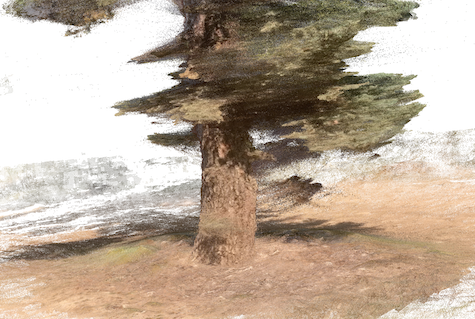}
	\caption{Tree 3D structure view-5}
	\label{fig:ex1_nerf8}
\end{subfigure}
	\begin{subfigure}[b]{0.6\textwidth}
	\centering
	\includegraphics[width=\textwidth]{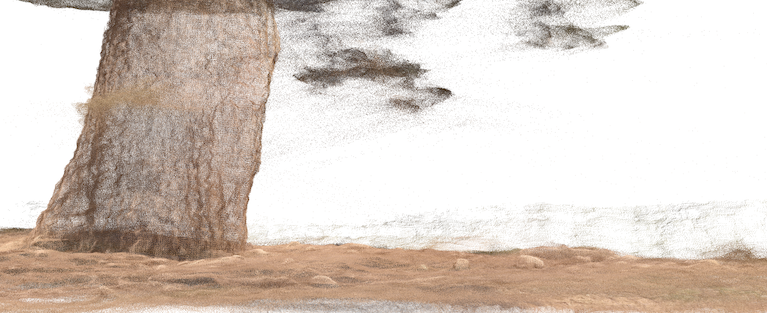}
	\caption{Fine 3D resolution of forest floor structure}
	\label{fig:ex1_nerf9}
\end{subfigure}

	\begin{subfigure}[b]{0.46\textwidth}
	\centering
	\includegraphics[width=\textwidth]{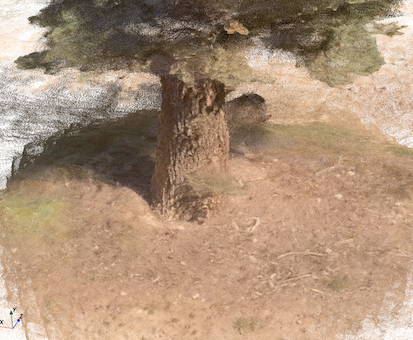}
	\caption{Forest floor 3D structure}
	\label{fig:ex1_nerf5}
\end{subfigure}
\begin{subfigure}[b]{0.254\textwidth}
	\centering
	\includegraphics[width=\textwidth]{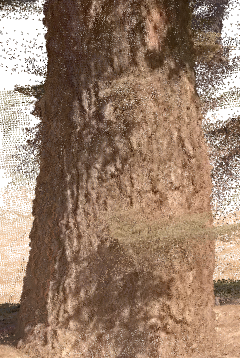}
	\caption{Tree trunk view-1}
	\label{fig:ex1_nerf6}
\end{subfigure}
\begin{subfigure}[b]{0.265\textwidth}
	\centering
	\includegraphics[width=\textwidth]{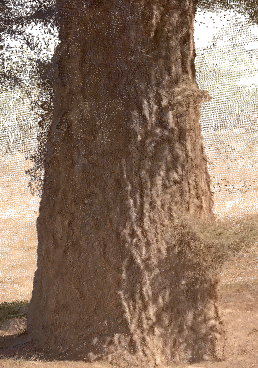}
	\caption{Tree trunk view-2}
	\label{fig:ex1_nerf7}
\end{subfigure}
	\caption{Neural field models are capable of extracting fine 3D structure from terrestrial multi-view images in forestry. Reconstructions demonstrate their potential to represent fine scale variability in heterogeneous forest ecosystems.}
	\label{fig:ex1_nerf}
\end{figure}

Expanding on the findings of neural fields in man-made environments, we conducted additional experiments to demonstrate its effectiveness in representing fine 3D structure details in forest ecosystems.
Figs. \ref{fig:it_colmap} and \ref{fig:ex1_nerf}  shows the extracted 3D structure of a Ponderosa pine tree in New Mexico, captured using standard 12-megapixel camera phone images collected along an elliptical trajectory around the tree. Fig. \ref{fig:ex1_view1} shows a few of the input example terrestrial multi-view RGB images collected. Figs. \ref{fig:ex1_colmap1} and\ref{fig:ex1_colmap2} presents the image snapshot trajectory represented as red rectangles, along with two 3D structure views derived from a traditional structure from motion (SFM) method \cite{schonberger:2016} applied to the multi-view input images. Note that the level of spatial variability detail provided by this SFM method is significantly low considering the resolution provided by the set of input images. 

 Can the representational power of modern AI models do better than classical 3D structure extraction methods in Forestry? We extract 3D structure by neural fields using the same input images and obtain the result shown in Fig.\ref{fig:ex1_nerf}. Note that much finer spatial variability details can be resolved across the 3D structure including the ground, trunk, branches, leaves. Even fine woody debris as shown in Figs\ref{fig:ex1_nerf9}-.\ref{fig:ex1_nerf5} and, bark can be resolved as shown in Figs.\ref{fig:ex1_nerf6}-\ref{fig:ex1_nerf7} in contrast to the result of traditional SFM in Fig.\ref{fig:it_colmap}. Note that even points coming from images degraded by sun-glare as shown in Fig.\ref{fig:ex1_view1} landed in the tree within reasonable distances as shown in Fig.\ref{fig:ex_nerf1}, this is significant specially considering the severity of the glare effects present in the 2D RGB images. In general, terrestrial multi-view imagery based NERF can be used to extract fine 3D spatial resolution along the vertical dimension of a tree stand with a level of detail similar to TLS and with the additional advantage of providing color for every 3D point estimate. 

\section{Neural Radiance Fields: A framework for remote sensing fusion in forestry}
\label{Sec:fusion}

Neural fields, have also demonstrated their ability to provide representations suitable for combining data from multiple sensing modalities in as long as these are co-registered or aligned. The neural fields framework, which extracts 3D structure from multi-view images, enables direct fusion of information with 3D point cloud sources through point cloud prior constraints \cite{Roessle2022}. Here, we consider the case of fusing multi-view images from an RGB camera and point clouds from LiDAR. The difficulty in fusing camera and LiDAR information is that camera measures color radiance while LiDAR measures distance \cite{castorena2020motion}. Fortunately, the framework of neural radiance fields can be used to extract 3D structure from images thus enabling direct fusion of information from LiDAR. This can be done though a learning function that extracts a 3D structure promoting consistency between the multi-view images as leveraged by standard NERF \cite{mildenhall:2020} subject to LiDAR point cloud priors \cite{Roessle2022} as:
\begin{equation} \label{camer_lidar_loss}
	\L(\Th) = \underbrace{ \sum_{\r \in \R} \left [  \|  C( \r) - \hat{C} (\r, \Th)\|_{\ell_2}^2\right]}_{\L_{C}(\Th)} + \lambda \underbrace{ \sum_{\r \in \R} \left [  \|  z( \r) - \hat{z} (\r, \Th)\|_{\ell_2}^2\right]}_{\L_{D}(\Th)}
\end{equation}
where the first term $\L_{C}(\Th)$ is the standard NERF learning function promoting a 3D structure with consistency between image views while the second term $\L_{D}(\Th)$ enforces the LiDAR point cloud priors with $\hat{z} (\r, \Th)$ given as in Eq.\eqref{depth}. The benefit of imposing point cloud priors into neural fields is two-fold: (1) it enables expressing relative distances obtained from standard 3D reconstruction of multi-view 2D images in terms of real metrics (e.g., meters), and (2) neural fields tend to face challenges in accurately estimating 3D structures at high distances (typically in the order of several tens of meters), where the LiDAR point cloud priors can serve as a supervisory signal to guide accurate estimation, especially at greater distances. This can be beneficial, as distances in aerial imagery are generally distributed around large distances, which may pose challenges for 3D structure extraction methods.  
%

\subsection{Filing in the missing below-canopy structure in  ALS data with TLS}
\label{Ssec:ALS2TLS}
\begin{figure}[h]
	\centering
	\begin{subfigure}[b]{0.4\textwidth}
		\centering
		\includegraphics[width=\textwidth]{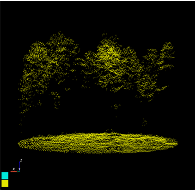}
		\caption{ALS side-view}
		\label{fig:ex1_als}
	\end{subfigure} 
	\hspace{1em}
	\begin{subfigure}[b]{0.4\textwidth}
		\centering
		\includegraphics[width=\textwidth]{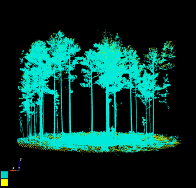}
		\caption{TLS side-view}
		\label{fig:ex2_tls}
	\end{subfigure}
	\caption{Forest structure from TLS and ALS: ALS provides sparse spatial information and is not capable of resolving sub-canopy detail. TLS on the other hand, provides fine spatial variability and resolution along full 3D vertical stands. }
	\label{fig:differences}
\end{figure}

\textit{In-situ} terrestrial laser scanning (TLS) has been demonstrated as a powerful tool for rapid assessment of forest structure in ecosystem monitoring and characterization. It is capable of very fine resolution including the vertical direction: surface, sub-canopy and canopy structure. However, its utility and application is restricted by limited spatial coverage. Aerial laser scanning (ALS) on the other hand, has the ability to rapidly survey broad scale areas at the landscape level, but is limited as it sparsely samples the scene providing only coarse spatial variability details and it also cannot penetrate the tree canopy. Fig. \ref{fig:ex1_als} shows a point cloud example collected using a full-waveform ALS system which collects $\approx$ 10 points per meter square. In Fig. \ref{fig:ex1_als}  note that the sub-canopy structure is not spatially resolved. In contrast, TLS is finely resolved below the canopy as observed in Fig.\ref{fig:ex2_tls}. 
%

\begin{figure}[h]
	\centering
	\begin{subfigure}[b]{0.447\textwidth}
		\centering
		\includegraphics[width=\textwidth]{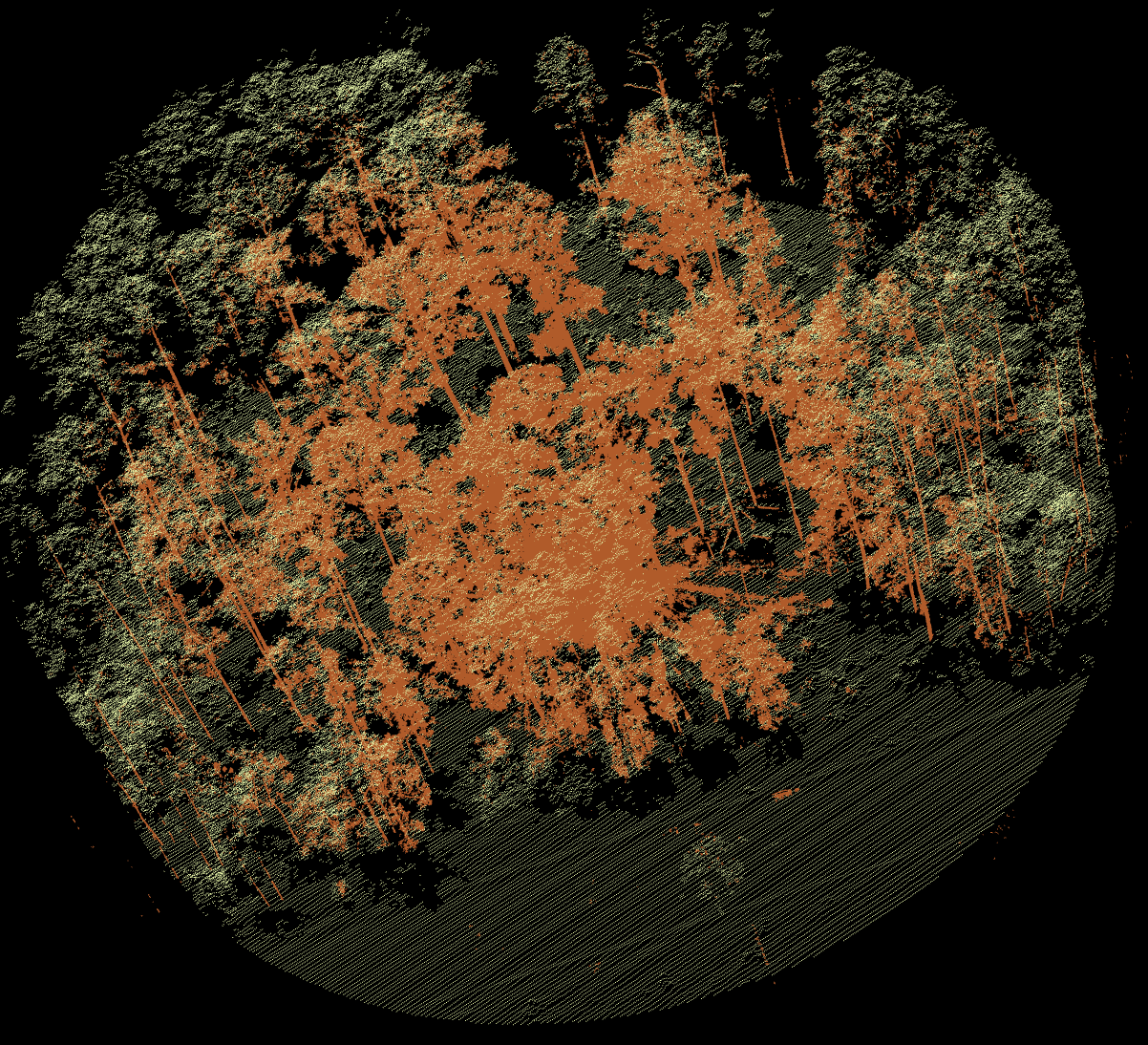}
		\caption{Co-registration Example 1}
		\label{fig:ex1}
	\end{subfigure} 
	\hspace{1em}
	\begin{subfigure}[b]{0.4\textwidth}
		\centering
		\includegraphics[width=\textwidth]{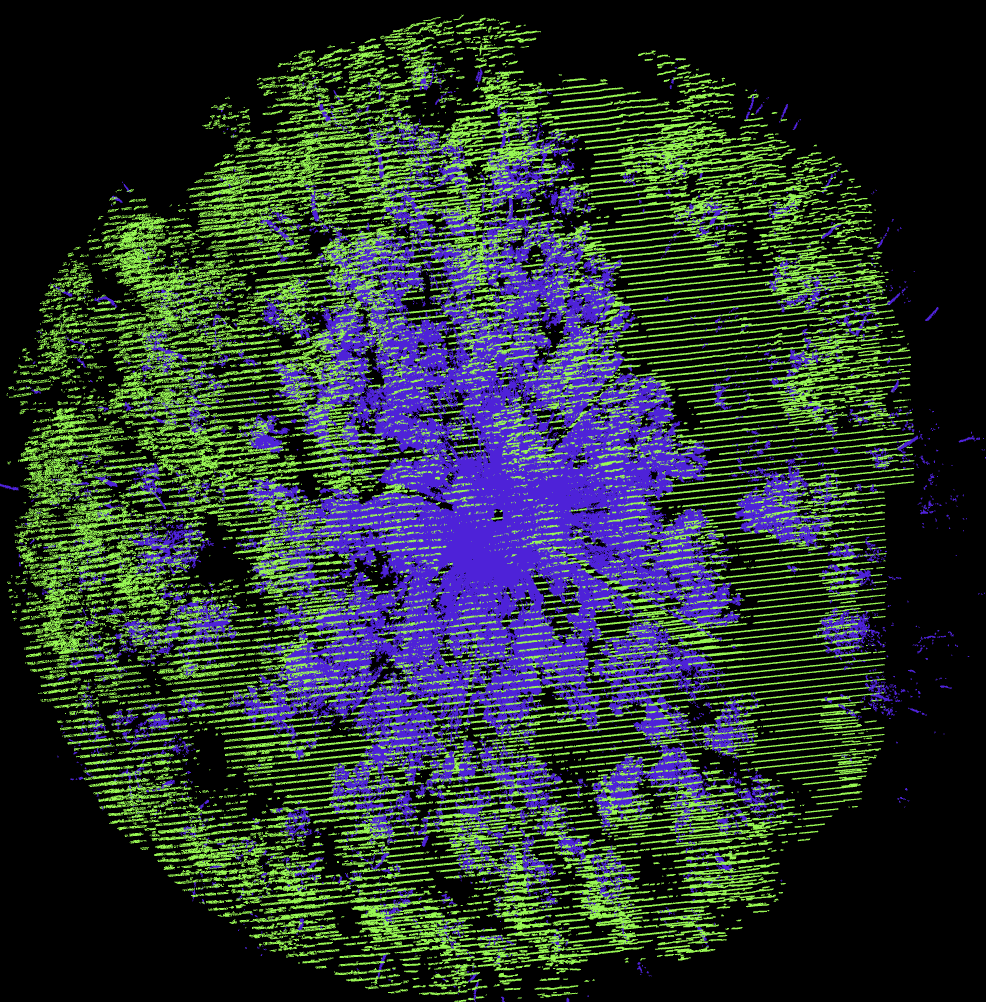}
		\caption{Co-registration Example 2}
		\label{fig:ex2}
	\end{subfigure}
	\caption{TLS to ALS co-registration: Forest features are well aligned qualitatively between both ALS and TLS sensing.}
	\label{fig:coregistration}
\end{figure}

Fortunately, the drawbacks of TLS and ALS scans can be resolved by co-registration which transforms the data to enable direct fusion. Here, we use the automatic and targetless based approach of \cite{castorena2023automated}. This was demonstrated to outperform standard methods \cite{Besl:1992}, \cite{myronenko2010point}, \cite{gao2019filterreg} in natural ecosystems and to be robust to resolution scales, view-points, scan area overlap, vegetation heterogeneity, topography and to ecosystem changes induced by pre/post low-intensity fire effects. It is also fully automatic, capable of self-correcting in cases of noisy GPS measurements and does not require any manually placed targets \cite{ge2021target} while performing at the same levels of accuracy. All TLS scans where co-registered into the coordinate system of ALS. Once scans have been co-registered they can be projected into a common coordinate system. Illustrative example results for two forest plots where included in Fig.\ref{fig:coregistration} where the two sources: ALS and TLS have been color coded differently, with the sparser point cloud being that of the ALS. Throughout all cases the co-registration produced finely aligned point clouds.
In general, the error produced by this co-registration method is $<$6 cm for the translation and $<$0.1$^{\text{o}}$ for the rotation parameters. The translation error in mainly due to the resolution of ALS at 10 points/meter square.

\subsection{Aerial Imagery} 
\begin{figure}[h]
	\centering
	\begin{minipage}{0.43\linewidth}
	\begin{subfigure}[b]{0.49\textwidth}
		\centering
		\includegraphics[width=\textwidth]{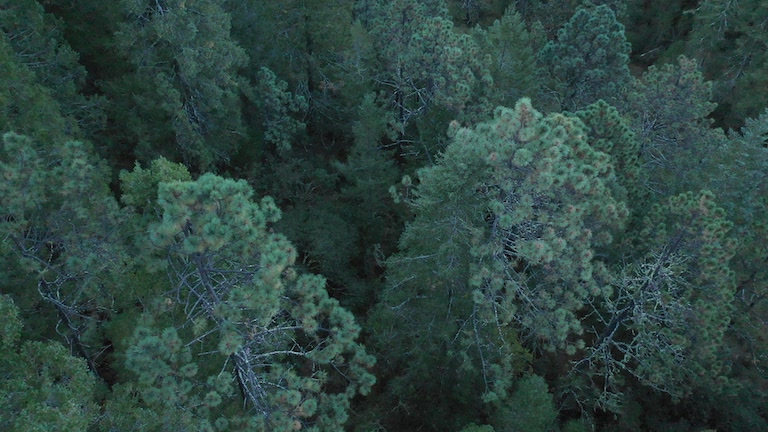}
		\caption{Image view-1}
		\label{fig:ex2_view1}
	\end{subfigure}
	\begin{subfigure}[b]{0.49\textwidth}
		\centering
		\includegraphics[width=\textwidth]{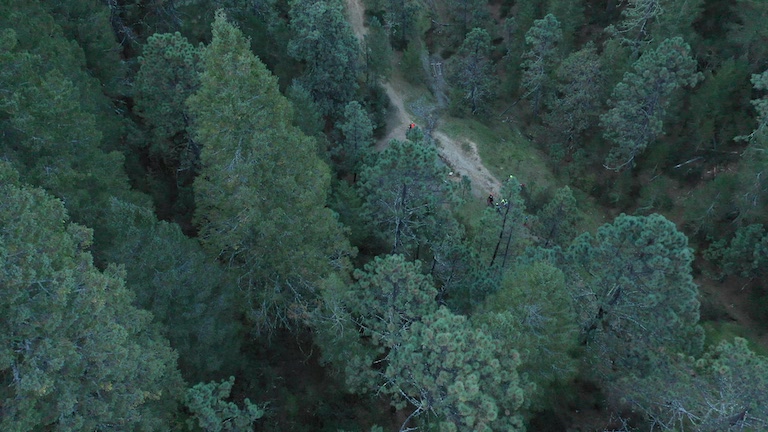}
		\caption{Image view-2}
		\label{fig:ex2_view2}
	\end{subfigure}

	\begin{subfigure}[b]{0.49\textwidth}
		\centering
		\includegraphics[width=\textwidth]{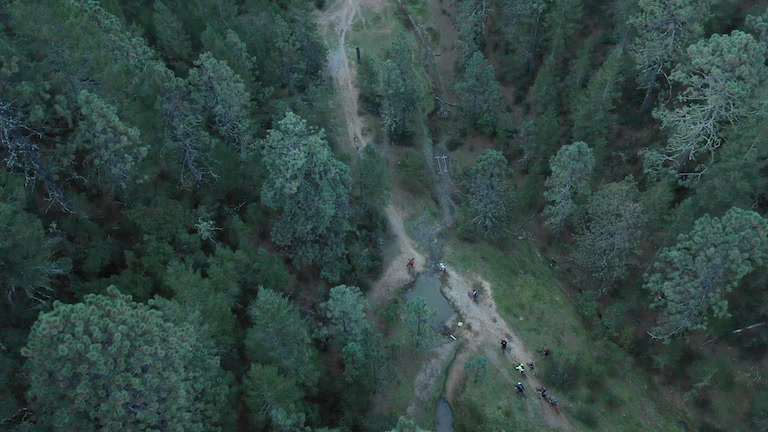}
		\caption{Image view-3}
		\label{fig:ex2_view3}
	\end{subfigure}	
	\begin{subfigure}[b]{0.49\textwidth}
		\centering
		\includegraphics[width=\textwidth]{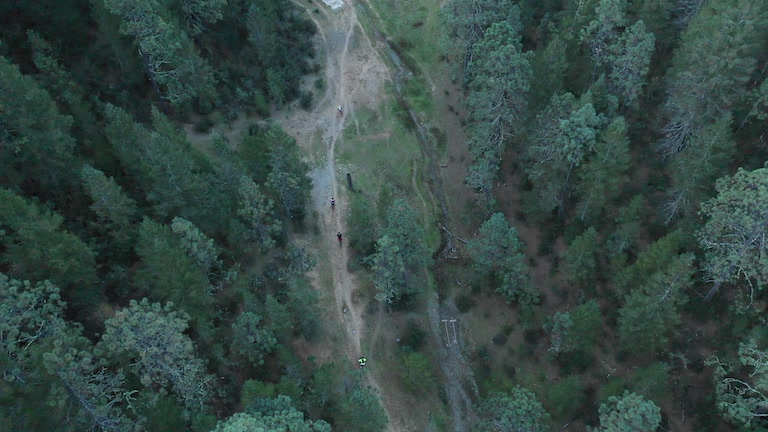}
		\caption{Image view-4}
		\label{fig:ex2_view4}
	\end{subfigure}

	\begin{subfigure}[b]{0.49\textwidth}
		\centering
		\includegraphics[width=\textwidth]{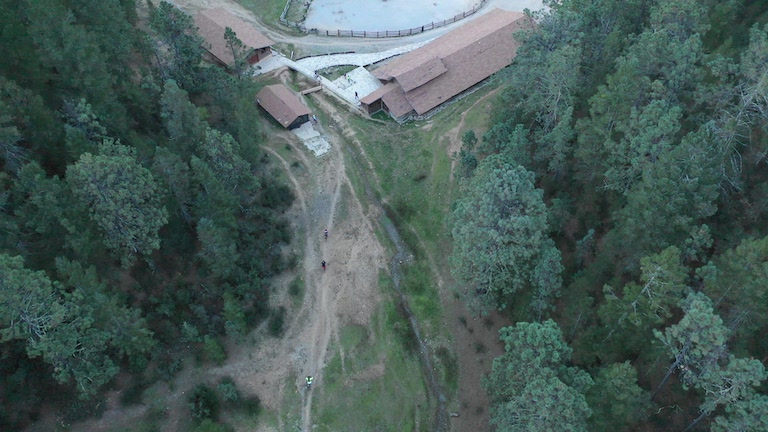}
		\caption{Image view-5}
		\label{fig:ex2_view5}
	\end{subfigure}
	\begin{subfigure}[b]{0.49\textwidth}
		\centering
		\includegraphics[width=\textwidth]{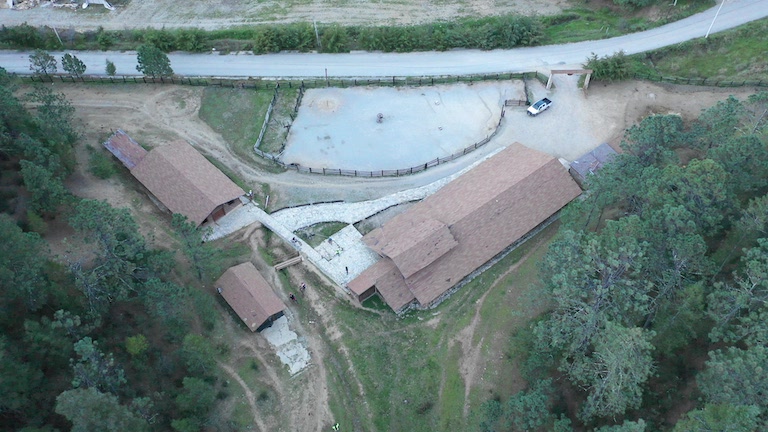}
		\caption{Image view-6}
		\label{fig:ex2_view6}
	\end{subfigure}
    \end{minipage}
    \hfill
    \begin{minipage}{0.252\linewidth}
        \begin{subfigure}{\textwidth}
	   \centering
	   \includegraphics[width=\textwidth]{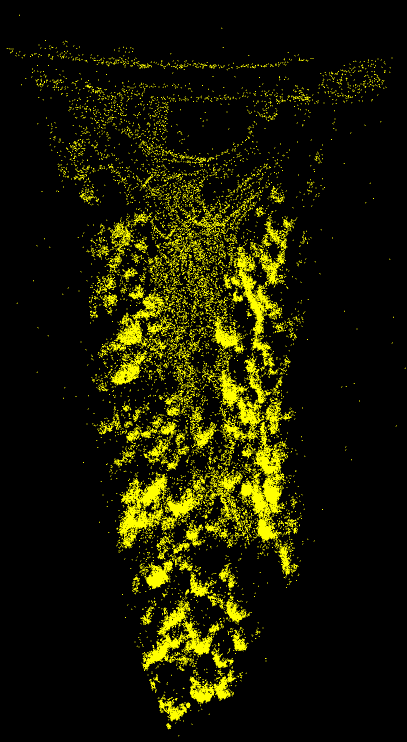}
	   \caption{Point Cloud}
	   \label{fig:ex2_point_cloud}
        \end{subfigure}
    \end{minipage}
    \hfill
    \begin{minipage}{0.28\linewidth}
        \begin{subfigure}{\textwidth}
	   \centering
	   \includegraphics[width=\textwidth]{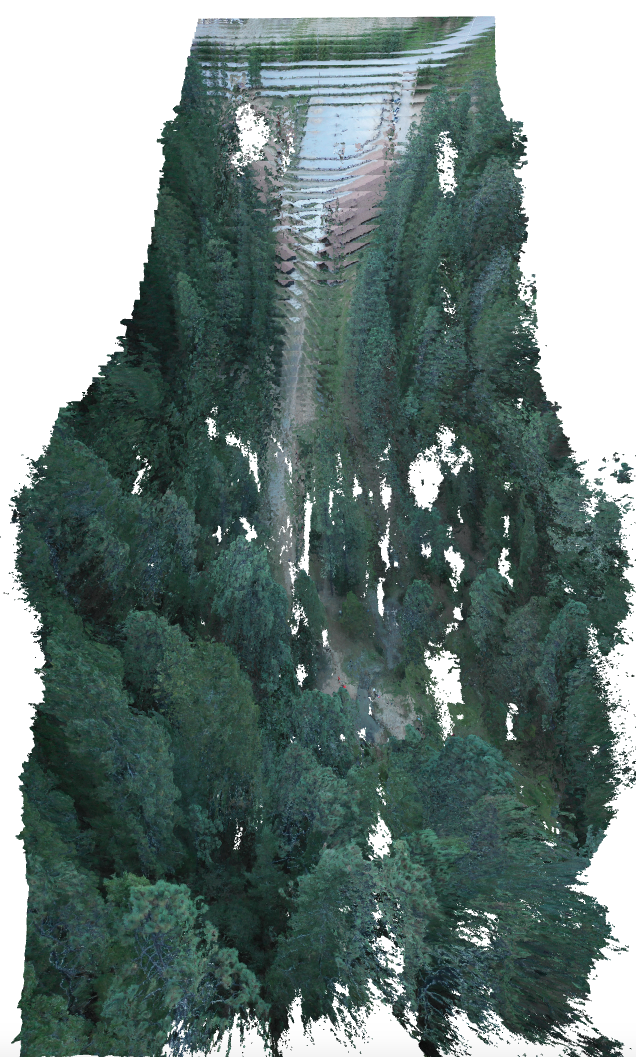}
	   \caption{NERF artifacts}
	   \label{fig:nerf_artifacts}
        \end{subfigure}
    \end{minipage}

	\begin{subfigure}[b]{0.32\textwidth}
		\centering
		\includegraphics[width=\textwidth]{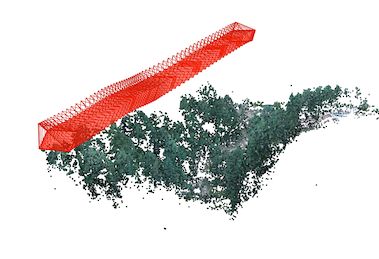}
		\caption{COLMAP view-1}
		\label{fig:ex2_colmap1}
	\end{subfigure}
	\begin{subfigure}[b]{0.32\textwidth}
		\centering
		\includegraphics[width=\textwidth]{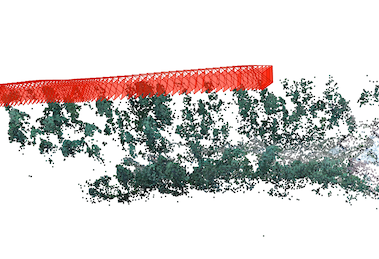}
		\caption{COLMAP view-2}
		\label{fig:ex2_colmap2}
	\end{subfigure}
	\begin{subfigure}[b]{0.32\textwidth}
		\centering
		\includegraphics[width=\textwidth]{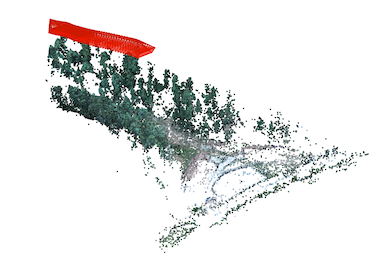}
		\caption{COLMAP view-3}
		\label{fig:ex2_colmap3}
	\end{subfigure}

     \begin{minipage}{0.5\linewidth}

        \begin{subfigure}{\linewidth}
	   \centering
	   \includegraphics[width=\textwidth]{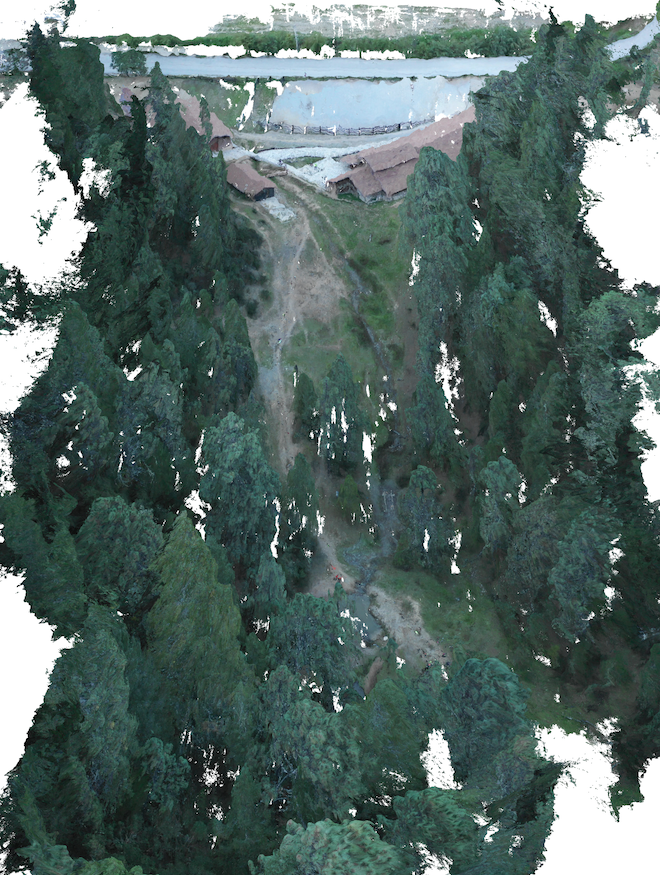}
	   \caption{NERF+LIDAR view-1}
	   \label{fig:ex2_nerf1}
        \end{subfigure}
    \end{minipage}
\hfill
\begin{minipage}{0.375\linewidth}
	\begin{subfigure}{\linewidth}
		\includegraphics[width=\textwidth]{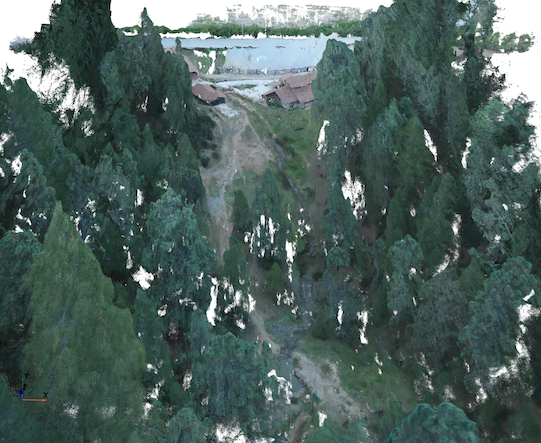}
		\caption{NERF+LIDAR view-2}
		\label{fig:ex2_nerf2}
	\end{subfigure}
	
	\begin{subfigure}{\linewidth}
		\centering
		\includegraphics[width=\textwidth]{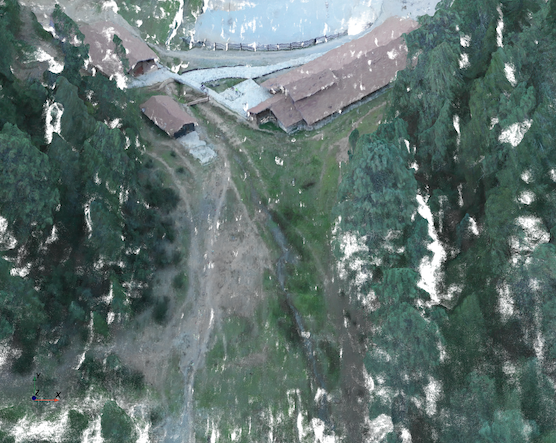}
		\caption{NERF+LIDAR view-3}
		\label{fig:ex2_nerf3}
	\end{subfigure}
\end{minipage}
	
	\caption{AI-based extraction of 3D structures from aerial multi-view 2D images + 3D point cloud data inputs. Imposing point cloud priors into 3D structure extraction improves distance ambiguities in structure and resolves artifact issues likely at far ranges.}
	\label{fig:drone_colmap}
\end{figure}

Experiments performed on broader forest areas were also conducted. Aerial RGB imagery was collected with a DJI Mavic2 Pro drone at 30Hz and a 3840 $\times$ 2160 pixel resolution. Figs. \ref{fig:ex2_view1}-\ref{fig:ex2_view6} show examples of multi-view aerial image inputs used by the SFM and neural fields models. The forest 3D structure resulting from running conventional SFM \cite{schonberger:2016} on these images is in Figs.\ref{fig:ex2_colmap1}-\ref{fig:ex2_colmap3} illustrating different perspective views. Again, the sequence of rectangles in red illustrate the drone flight path and the snapshot image locations. Note that SFM was capable of resolving 3D structure for the entire scene. 

Applying NERF directly into the RGB imagery dataset, did not result in comparable performance as in the case of the Ponderosa pine tree shown in Section \ref{Ssec:individual_tree}. Without point cloud constraints, the 3D structure extracted by the neural fields in Fig. \ref{fig:nerf_artifacts} shows the presence of artifacts at large distances. 
 The main reason for these artifacts is that NERF had difficulties in recovering 3D structures from images with objects distributed at far distances (e.g., ground surface in aerial scanning). Imposing LiDAR point cloud priors we hypothesize can help to alleviate this issue. Here, we follow the methodology of \cite{Roessle2022} and conduct experiments for fusing camera and LiDAR information through the learning function in Eq.\eqref{camer_lidar_loss}. The LiDAR point cloud uses both co-registered TLS and ALS data which provides information to constrain both distances in the mid-story below the canopy and those between the ground surface and the tree canopy. The co-registration approach used to align ALS and TLS point clouds is the one described in Section \ref{Ssec:ALS2TLS}. Note that TLS information is not available throughout the entire tested forest area; rather, only one TLS scan was collected. We found the information provided by just one single scan was enough to constraint the relative distances in sub-canopy areas throughout the entire scene.
 Imposing additional constraints through consistency with the input point cloud shown in Fig.\ref{fig:ex2_point_cloud}, results in the extracted 3D structure shown in Figs. \ref{fig:ex2_nerf1}-\ref{fig:ex2_nerf3}. In this case, the point cloud prior imposes constraints that resolve the associated difficulties at large distances. Note that this reconstruction is significantly less sparser than those shown in Figs.\ref{fig:ex2_colmap1}-\ref{fig:ex2_colmap2} obtained from conventional SFM. NERF+LIDAR results in improved resolution which in turn enables the detection of a much finer spatial variability, specially important for current existing demands in forest monitoring at broad scale. This illustrates the capacity of neural fields models not only to represent highly detailed 3D forest structure from aerial multi-view data but also the possibility of combining multi-source remotely sensed data (i.e., imagery and LiDAR).
%
%
%


\section{Prediction of forest factor metrics}
\label{Sec:prediction}
Demonstration of the described capabilities of neural fields on forest monitoring programs consists here in performance evaluations of 3D forest structure derived metrics. These can include for example number of trees, species composition, tree height, diameter at breast height (DBH), age on a given geo-referenced area. However, since our focus is to demonstrate the usefulness of neural radiance fields for representing 3D forest structure, we only illustrate its potential in prediction of the number of trees and DBH along geo-referenced areas. 
The data used includes overlapping TLS+ALS+GPS+aerial imagery multi-view multi-modal data collected over forest plot units. Each of these plots represents a location area of a varying size: some of size 20 x 50 m and others at 15 m radius. The sites in which data was collected is in northern New Mexico, USA (the NM dataset). 
The vegetation heterogeneity and topography variability of the landscape is significantly diverse. The NM site contains high elevation ponderosa pine and mixed-conifer forest: white fir, limber pine, aspen, Douglas fir and Gambel oak and topography is at high elevation and of high-variation (between 5,000-10,200 ft). 
The TLS data was collected using a LiDAR sensor mounted on a static tripod placed at the center of each plot. The ALS data was collected by a Galaxy T2000 LiDAR sensor mounted on a fixed-wing aircraft. The number of LiDAR point returns per volume depend on the sensor and scanning protocol settings (e.g., TLS or ALS, range distribution, number of scans) and these vary across plots depending on the heterogeneity of the site. 
\begin{figure} [h]
		\centering
	\begin{subfigure}[b]{0.7\textwidth}
		\centering
		\includegraphics[width=\textwidth]{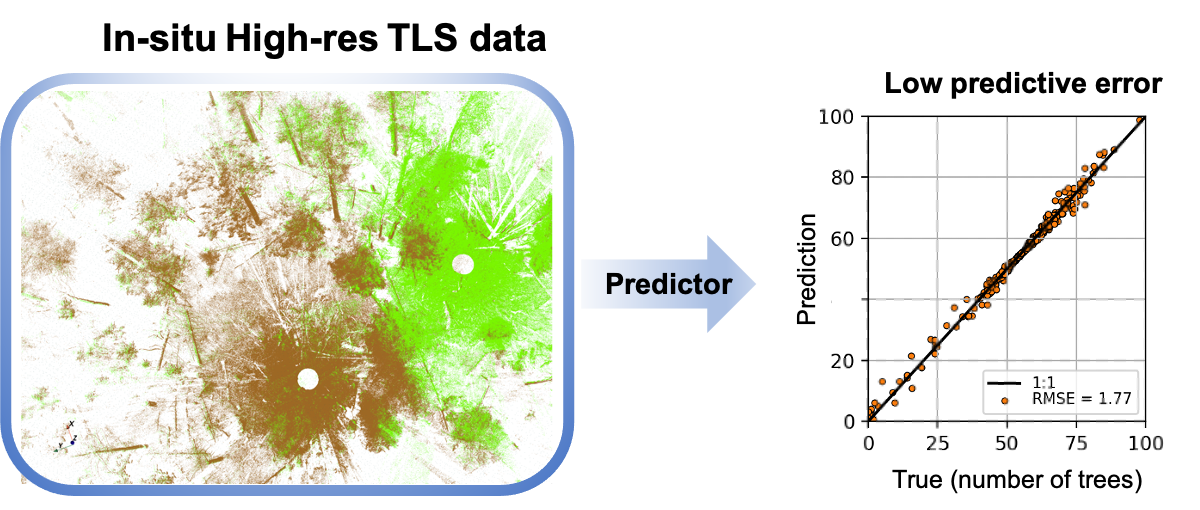}
		\caption{In-situ plot-scale TLS has demonstrated to be an effective tool in estimating plot-level vegetation characteristics }
		\label{fig:tls}
	\end{subfigure} 
	\hspace{1em}
	\begin{subfigure}[b]{0.7\textwidth}
		\centering
		\includegraphics[width=\textwidth]{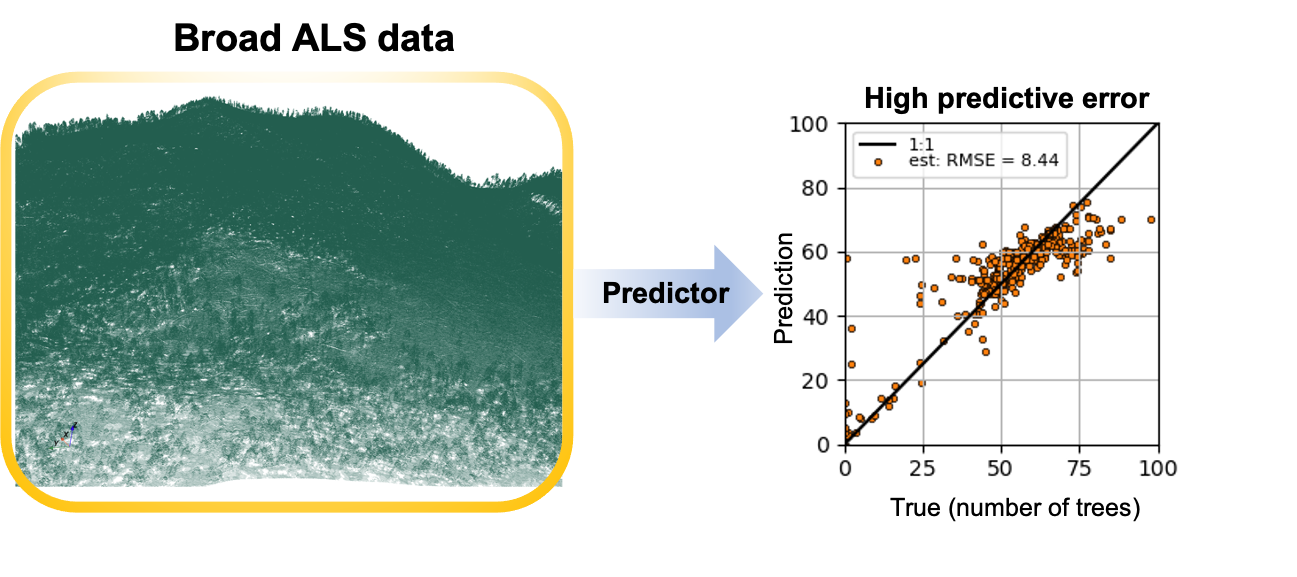}
		\caption{Broad-landscape scale ALS derived prediction, does not have vertical dimension resolution resulting in underestimate predictions }
		\label{fig:als}
	\end{subfigure}
	\caption{ LiDAR derived vegetation attribute estimation for single TLS and ALS.}
	\label{fig:als_vs_tls} 
\end{figure}
Ground truth number of trees per plot was obtained through standard forest plot field surveying techniques involving actual physical measurements of live/dead vegetation composition. 
Data from a total of 250 plots where collected in the NM dataset.
In every forest plot overlapping ALS, GPS, TLS and multi-view aerial imagery data was collected along with the corresponding field measured ground truth.
Prediction of the number of trees $y_1$ per plot given point cloud $\X$, was performed following the approach of the GRNet \cite{xie2020grnet} \cite{windrim2020detection}. In general, the methodology consists in computing 2D bounding boxes each corresponding to a tree detection from a birds eye view (BEV). A refinement segmentation approach then follows which projects each 2D bounding box into 3D space. The resulting points inside each 3D bounding box are then segmented by foliage, upper stem and lower stem and empty space and this information is used to improve estimates over the number of trees. This methodology is used independently on several case scenarios comparing the performance of a combination of remote sensing approaches: (1) neural fields (NF) from aerial RGB Images, (2) ALS as in Fig.\ref{fig:als}, (3) TLS as in Fig.\ref{fig:tls}, (4) ALS+TLS, (5) NF-RGB images + ALS, (6) NF-RGB images + TLS, (7) NF-RGB Images + TLS + ALS. Note that the TLS, ALS and TLS+ALS prediction results does not make any use of neural fields. Rather, their performance was included only for comparison purposes. Table \ref{tab:table1} summarizes the root mean squared error (RMSE) results for each of the tested cases.

\begin{table}[h]
	\centering
	\caption{RMSE Prediction performance of number of trees per plot in NM dataset.}\label{tab:table1}
	\begin{tabular}{l|c|c|c|c|c|c|c|}
		\toprule 
		\bfseries Method & NF-RGB & ALS & TLS & ALS+TLS & NF-RGB+ALS & NF-RGB+TLS & NF-RGB+ALS+TLS \\
		\midrule 
		\bfseries RMSE & 10.61 & 8.44 & 1.77  & 1.67 & 1.41 & 1.39 & 1.32 \\
		\bottomrule 
	\end{tabular}
\end{table}
The results in Table \ref{tab:table1} corroborate some of the trade-offs between the sensing modalities and in addition some of the advantages gained through the use of neural fields in forestry. First, the superiority of TLS over ALS data on the number of trees metric is mainly due to the presence of information in sub-canopy which is characteristic of \text{in-situ} TLS. This in alignment with current demonstrations in the literature which have motivated the widespread usage of in-situ TLS in forestry applications even though it is not as spatially scalable as ALS is \cite{pokswinski2021simplified}. We would have seen the opposite relationships between TLS and ALS, however, in cases when the plot size is significantly higher than the range of a single in-situ TLS scan. A problem which can be resolved by adding multiple view co-registered TLS scans per plot. This limitation is caused as the sensor remains static at collection time which makes it more susceptible to occlusions, specially in dense forest areas where trees can significantly reduce the view of TLS at higher ranges. TLS+ALS overcomes, on the other hand,  the limitations of the individual LiDAR platforms by filling in the missing information characteristic of each platform. Structure from neural fields using only multi-view RGB images performed slightly worst than both ALS and TLS. This may be due to the limited number of multi-view images collected per plot, the performance for deriving structure from NERF or to the joint performance of NERF in conjunction with the GRNet. Fortunately, fusing neural fields from multi-view imagery with LiDAR shows a significant improvement overall fused cases (i.e., NF+ALS, NF+TLS and NF+ALS+TLS). We see that the prior supervisory signal imposed by the LiDAR point cloud helps on guiding the resulting 3D structure from NERF to alleviate the artifacts arising at far distances when using multi-view imagery only. We would like to finalize this discussion by highlighting the performance of the NF-RGB+ALS method which is marginally similar to the best performing method (i.e., NF-RGB+ALS+TLS). The benefit of using NF-RGB+ALS is that being both airborne makes the data collection of these two modalities time and cost efficient, in contrast, to in-situ remote sensing methods such as TLS. This has significant implications towards achieving both scalable and highly performing forest monitoring programs. In general, one has to resort to a balance between scalability and performance performance depending on needs. Our work instead, offers a method which can potentially achieve similar performance as in-situ methods with the benefits of scalability over the landscape scales through computational methods.

Additional experiments were conducted to explore the ability of neural fields from terrestrial based multi-view imagery to achieve a performance near that of TLS in metrics that depend on sub-canopy information. In this case, we evaluated performance on the DBH metric for a total of 200 trees. Ground truth DBH was manually measured in the field for each tree's stem diameter at a height of 1.3m. A total of 5 co-registered TLS scans where used per tree, each collected from a different location and viewing each tree from a different perspective to reduce the effects of occlusion and to remove the degrading effects of lower point LiDAR return densities at farther ranges. Multi-view TLS co-registration was obtained using the method of \cite{castorena2023automated}. Terrestrial multi-view RGB imagery data for NERF was collected around an oblique trajectory around each tree as exemplified in Fig.\ref{fig:it_colmap} with $10-15$ snapshot images per tree. Algorithmic performance for estimating DBH was compared against TLS, ALS, TLS+ALS and NF-RGB. The estimation approach of \cite{windrim2020detection} relying on stem geometric circular shape fitting at a height of 1.3m over the ground was used following their implementation. Performance is measured as the average error as a percentage of the actual field measured DBH ground truth, following the work of \cite{windrim2020detection}. Comparison results are reported in Table \ref{tab:table2}. 
\begin{table}[h]
	\centering
	\caption{Comparison of sensing modalities on average error DBH estimation.}\label{tab:table2}
	\begin{tabular}{l|c|c|c|c|}
		\toprule 
		\bfseries Method & NF-RGB & ALS & TLS & ALS+TLS \\
		\midrule 
		\bfseries Avg. error $\%$ & 1.7 $\%$ & 32.7$\%$ & 1.3$\%$ & 3.3$\%$ \\
		\bottomrule 
	\end{tabular}
\end{table}

In table \ref{tab:table2} ALS performs the worst DBH estimation due to its inherited limited sub-canopy resolution. Multi-view TLS on the other hand, performs the best at 1.3$\%$ error consistent with TLS superiority findings in \cite{windrim2020detection} for metrics relying on sub-canopy information. However, our neural fields approach from terrestrial imagery performs marginally on par with multi-view TLS, with the additional advantage that RGB camera sensors are simpler to access commercially and significantly cheaper than LiDAR.

In terms of computational specifications, neural radiance fields were trained using a set of overlapping 10-50 multi-view images per scene. The fast implementation of \cite{mueller2022instant} was used with training on the terrestrial and aerial multi-view imagery taking from 30-60 secs per 3D structure extraction (e.g., per plot in the aerial imagery case, per tree in the terrestrial imagery case). Adding the LiDAR constraints was done following the implementation from \cite{Roessle2022}. The neural radiance architecture is a multilayer perceptron (MLP) with two hidden layers and a ReLU layer per hidden layer and a linear output layer as in \cite{mueller2022instant}. Training was performed using the ADAM optimizer \cite{kingma2014adam} with parameters $\beta_1$ = 0.9, $\beta_2$ = 0.99, $\epsilon =10^{-15}$ using NVIDIA Tesla V100.

The main limitation of neural fields from aerial multi-view imagery is the presence of occlusion of sub-canopy structure, specially in densely forested areas. In our case, fusion with TLS data can resolve this problem as terrestrial data provides highly detailed sub-canopy information. Additionally, when TLS is unavailable, terrestrial imagery can be used instead. Our 3D structure experiments from terrestrial multi-view information in Sec.\ref{Ssec:individual_tree} and the DBH estimation performance results demonstrate that highly detailed structure along the entire vertical stand direction can be extracted by neural fields when image information is available. In the absence of multi-view image data, however, neural fields are not capable of generating synthetic information behind occluded areas and performance on metrics affected by occlusion are expected to yield large errors. This problem can be alleviated through multi-view images capturing the desired areas of interest in the ecosystem.

\section{Conclusion}
\label{Sec:conclusion}
In this work, we proposed neural radiance fields as representations that can finely express the 3D structure of forests both in the \textit{in-situ} and at the broad landscape scale. In addition, the properties of neural radiance fields; in particular, the fact that they account for both the origin and direction of radiance to define 3D structure enables the fusion of data coming from multiple locations and modalities; more specifically those from multi-view LiDAR's and cameras. Finally, we evaluated the performance of 3D structure derived metrics typically used in forest monitoring programs and demonstrated the potential of neural fields to improve performance of scalable methods at near the level of \textit{in-situ} methods. This not only represents a benefit on sampling time efficiency but also has powerful implications on reducing  monitoring costs.

\section*{Acknowledgements}
Research presented in this article was supported by the Laboratory Directed Research and Development program of Los Alamos National Laboratory under project number GRR0CSRN.



\begin{thebibliography}{10}
\providecommand{\url}[1]{\texttt{#1}}
\providecommand{\urlprefix}{URL }
\providecommand{\doi}[1]{https://doi.org/#1}

\bibitem{Atchley2021}
Atchley, A., Linn, Rodman, J.A., Hoffman, C., Hyman, J.D., Pimont, F., Sieg,
  C., Middleton, R.S.: Effects of fuel spatial distribution on wildland fire
  behaviour. International Journal of Wildland Fire  \textbf{30}(3),  179--189
  (2021)

\bibitem{Besl:1992}
{Besl}, P.J., {McKay}, N.D.: A method for registration of 3-d shapes. IEEE
  Transactions on Pattern Analysis and Machine Intelligence  \textbf{14}(2),
  239--256 (Feb 1992). \doi{10.1109/34.121791}

\bibitem{castorena:2010}
Castorena, J., Creusere, C.D., Voelz, D.: Modeling lidar scene sparsity using
  compressive sensing. In: 2010 IEEE International Geoscience and Remote
  Sensing Symposium. pp. 2186--2189. IEEE (2010)

\bibitem{castorena2023automated}
Castorena, J., Dickman, L.T., Killebrew, A.J., Gattiker, J.R., Linn, R.,
  Loudermilk, E.L.: Automated structural-level alignment of multi-view tls and
  als point clouds in forestry (2023)

\bibitem{castorena2020motion}
Castorena, J., Puskorius, G.V., Pandey, G.: Motion guided lidar-camera
  self-calibration and accelerated depth upsampling for autonomous vehicles.
  Journal of Intelligent \& Robotic Systems  \textbf{100}(3),  1129--1138
  (2020)

\bibitem{dubayah2000lidar}
Dubayah, R.O., Drake, J.B.: Lidar remote sensing for forestry. Journal of
  forestry  \textbf{98}(6),  44--46 (2000)

\bibitem{FaoUnep2020}
FAO, U.: The state of the world's forests 2020. In: Forests, biodiversity and
  people. p.~214. Rome, Italy (2020). \doi{https://doi.org/10.4060/ca8642en}

\bibitem{gao2019filterreg}
Gao, W., Tedrake, R.: Filterreg: Robust and efficient probabilistic point-set
  registration using gaussian filter and twist parameterization. In:
  Proceedings of the IEEE/CVF Conference on Computer Vision and Pattern
  Recognition. pp. 11095--11104 (2019)

\bibitem{ge2021target}
Ge, X., Zhu, Q.: Target-based automated matching of multiple terrestrial laser
  scans for complex forest scenes. ISPRS Journal of Photogrammetry and Remote
  Sensing  \textbf{179},  1--13 (2021)

\bibitem{hilker2010comparing}
Hilker, T., van Leeuwen, M., Coops, N.C., Wulder, M.A., Newnham, G.J., Jupp,
  D.L., Culvenor, D.S.: Comparing canopy metrics derived from terrestrial and
  airborne laser scanning in a douglas-fir dominated forest stand. Trees
  \textbf{24}(5),  819--832 (2010)

\bibitem{hyyppa2012advances}
Hyypp{\"a}, J., Yu, X., Hyypp{\"a}, H., Vastaranta, M., Holopainen, M., Kukko,
  A., Kaartinen, H., Jaakkola, A., Vaaja, M., Koskinen, J., et~al.: Advances in
  forest inventory using airborne laser scanning. Remote sensing
  \textbf{4}(5),  1190--1207 (2012)

\bibitem{kajiya:1984}
Kajiya, J.T., Von~Herzen, B.P.: Ray tracing volume densities. ACM SIGGRAPH
  computer graphics  \textbf{18}(3),  165--174 (1984)

\bibitem{kankare2014estimation}
Kankare, V., Joensuu, M., Vauhkonen, J., Holopainen, M., Tanhuanp{\"a}{\"a},
  T., Vastaranta, M., Hyypp{\"a}, J., Hyypp{\"a}, H., Alho, P., Rikala, J.,
  et~al.: Estimation of the timber quality of scots pine with terrestrial laser
  scanning. Forests  \textbf{5}(8),  1879--1895 (2014)

\bibitem{kingma2014adam}
Kingma, D.P., Ba, J.: Adam: A method for stochastic optimization. arXiv
  preprint arXiv:1412.6980  (2014)

\bibitem{lausch2017understanding}
Lausch, A., Erasmi, S., King, D.J., Magdon, P., Heurich, M.: Understanding
  forest health with remote sensing-part ii—a review of approaches and data
  models. Remote Sensing  \textbf{9}(2), ~129 (2017)

\bibitem{linn2002}
Linn, R., Reisner, J., Colman, J.J., Winterkamp, J.: Studying wildfire behavior
  using firetec. International journal of wildland fire  \textbf{11(4)},
  233--246. (2002)

\bibitem{mildenhall:2020}
Mildenhall, B., Srinivasan, P.P., Tancik, M., Barron, J.T., Ramamoorthi, R.,
  Ng, R.: Nerf: Representing scenes as neural radiance fields for view
  synthesis. arXiv preprint arXiv:2003.08934  (2020)

\bibitem{mueller2022instant}
M\"uller, T., Evans, A., Schied, C., Keller, A.: Instant neural graphics
  primitives with a multiresolution hash encoding. ACM Trans. Graph.
  \textbf{41}(4),  102:1--102:15 (Jul 2022). \doi{10.1145/3528223.3530127},
  \url{https://doi.org/10.1145/3528223.3530127}

\bibitem{myronenko2010point}
Myronenko, A., Song, X.: Point set registration: Coherent point drift. IEEE
  transactions on pattern analysis and machine intelligence  \textbf{32}(12),
  2262--2275 (2010)

\bibitem{pokswinski2021simplified}
Pokswinski, S., Gallagher, M.R., Skowronski, N.S., Loudermilk, E.L., Hawley,
  C., Wallace, D., Everland, A., Wallace, J., Hiers, J.K.: A simplified and
  affordable approach to forest monitoring using single terrestrial laser scans
  and transect sampling. MethodsX  \textbf{8},  101484 (2021)

\bibitem{Roessle2022}
Roessle, B., Barron, J.T., Mildenhall, B., Srinivasan, P.P., Niebner, M.: Dense
  depth priors for neural radiance fields from sparse input views. In:
  Proceedings of the IEEE conference on computer vision and pattern
  recognition. pp. 12892--12901 (2022)

\bibitem{schonberger:2016}
Schonberger, J.L., Frahm, J.M.: Structure-from-motion revisited. In:
  Proceedings of the IEEE conference on computer vision and pattern
  recognition. pp. 4104--4113 (2016)

\bibitem{tomppo2010national}
Tomppo, E., Gschwantner, T., Lawrence, M., McRoberts, R.E., Gabler, K.,
  Schadauer, K., Vidal, C., Lanz, A., Staahl, G., Cienciala, E.: National
  forest inventories. Pathways for Common Reporting. European Science
  Foundation  \textbf{1},  541--553 (2010)

\bibitem{vierling2008lidar}
Vierling, K.T., Vierling, L.A., Gould, W.A., Martinuzzi, S., Clawges, R.M.:
  Lidar: shedding new light on habitat characterization and modeling. Frontiers
  in Ecology and the Environment  \textbf{6}(2),  90--98 (2008)

\bibitem{white2016remote}
White, J.C., Coops, N.C., Wulder, M.A., Vastaranta, M., Hilker, T., Tompalski,
  P.: Remote sensing technologies for enhancing forest inventories: A review.
  Canadian Journal of Remote Sensing  \textbf{42}(5),  619--641 (2016)

\bibitem{windrim2020detection}
Windrim, L., Bryson, M.: Detection, segmentation, and model fitting of
  individual tree stems from airborne laser scanning of forests using deep
  learning. Remote Sensing  \textbf{12}(9), ~1469 (2020)

\bibitem{xie2020grnet}
Xie, H., Yao, H., Zhou, S., Mao, J., Zhang, S., Sun, W.: Grnet: Gridding
  residual network for dense point cloud completion. In: ECCV (2020)

\end{thebibliography}

\appendix

\end{document}